\documentclass[11pt]{article}
\usepackage[margin=1in]{geometry}
\usepackage{amsmath,amssymb}
\usepackage{graphicx}
\usepackage{booktabs}
\usepackage{hyperref}
\usepackage{natbib}
\usepackage{xcolor}

\title{Continuous-Depth Transformers with Learned Control Dynamics}
\author{Peter Jemley\\
\texttt{jemley.p@northeastern.edu}}
\date{January 2026}

\begin{document}
\maketitle

\begin{abstract}
We present a hybrid transformer architecture that replaces discrete middle layers with a continuous-depth Neural Ordinary Differential Equation (ODE) block, enabling inference-time control over generation attributes via a learned steering signal. Unlike standard transformers that process representations through fixed discrete layers, our approach treats depth as a continuous variable governed by a learned vector field $F_\theta(H, \tau, u)$, where $u$ is a low-dimensional control signal injected via explicit concatenation. We validate the architecture through four experiments: (1) gradient flow stability with zero exploding/vanishing gradient events, (2) semantic steering achieving 98\%/88\% accuracy for positive/negative sentiment control, (3) continuous interpolation validated by a negligible 0.068\% trajectory divergence between fixed and adaptive solvers, and (4) efficiency benchmarking demonstrating latency parity with standard discrete baselines. Additionally, we show that adaptive ODE solvers reveal geometric structure in the learned dynamics: the control signal partitions the vector field into distinct dynamical regimes with different curvature characteristics. The adjoint method enables $O(1)$ memory training regardless of integration depth. Our results demonstrate that continuous-depth dynamics with learned control signals provide a viable, efficient mechanism for steerable language generation.
\end{abstract}

\section{Introduction}

Autoregressive transformers optimize for next-token prediction, producing locally probable continuations that are grammatically correct but often lack fine-grained controllability. Users seeking to shift generation along interpretable axes---creativity, formality, sentiment---must resort to prompt engineering, temperature adjustment, or rejection sampling, all of which are inefficient and unpredictable.

We propose treating the transformer's depth dimension as a continuous variable, replacing a subset of residual blocks with a neural ODE that admits an external control signal. This reframes generation as trajectory evolution through a learned dynamical system, where the control signal biases the trajectory without requiring discrete mode switches or weight updates.

\subsection{Contributions}

\begin{itemize}
    \item A hybrid architecture combining discrete transformer layers with continuous ODE dynamics, using a learned output scale $\alpha$ for stability and explicit control injection.
    \item Validated controllability: the control signal $u$ successfully steers sentiment with 98\%/88\% accuracy for positive/negative targets.
    \item Proposal of the ``Solver Invariance Test'': We formalize a Popperian diagnostic for continuous-depth architectures. By measuring the trajectory divergence between fixed-step training solvers and adaptive inference solvers, we provide a quantitative metric (0.068\% in our case) to falsify the hypothesis that a model has overfitted to discrete layers.
    \item Practical efficiency: Benchmarks demonstrate inference latency comparable to standard transformers, effectively eliminating the computational overhead typically associated with ODE integration.
    \item Interpretability via solver probing: Adaptive solvers reveal that positive and negative sentiment occupy geometrically distinct regions of the learned dynamics.
\end{itemize}

\section{Related Work}

\subsection{Neural ODEs and Continuous-Depth Networks}

\citet{chen2018neural} introduced neural ODEs, showing that residual networks can be viewed as Euler discretizations of continuous dynamics. Their formulation enables constant-memory training via the adjoint method and adaptive compute via variable solver steps. \citet{dupont2019augmented} addressed expressivity limitations by augmenting the state space.

Applied to transformers, \citet{li2022ode} proposed ODE Transformer, reinterpreting transformer blocks as Runge-Kutta discretizations and achieving state-of-the-art on WMT translation benchmarks. Our work differs in focus: we target inference-time controllability via an explicit control signal, rather than task performance or parameter efficiency.

\subsection{Activation Steering}

\citet{turner2023steering} introduced activation engineering, computing steering vectors from contrastive prompt pairs and adding them to residual streams at inference time. This achieves strong results on sentiment control and detoxification. \citet{zou2023representation} extended this to representation engineering for safety applications.

Our approach differs fundamentally: activation steering discovers steering vectors post-hoc from existing representations, while our control dynamics are learned end-to-end during training. Additionally, ODE-based control operates continuously across depth rather than at fixed injection points.

\section{Architecture}

\subsection{Continuous-Depth Flow Module}

Let $H(\tau) \in \mathbb{R}^{B \times T \times D}$ denote the hidden state at continuous depth $\tau \in [0, 1]$. We replace $k$ consecutive residual blocks with an Initial Value Problem (IVP):
\begin{align}
    \frac{dH}{d\tau} &= \alpha \cdot F_\theta(H, \tau, u) \label{eq:ode}\\
    H(1) &= H(0) + \int_0^1 \alpha \cdot F_\theta(H(t), t, u) \, dt \label{eq:solve}
\end{align}
where $F_\theta$ is a neural network parameterizing the vector field and $\alpha$ is a learned scaling factor initialized to 0.1 for stability.

To ensure the control signal $u \in \mathbb{R}^c$ ($c \ll D$) effectively biases the dynamics at every depth, we define $F_\theta$ via explicit concatenation:
\begin{equation}
    F_\theta(H, \tau, u) = \text{MLP}\left(\text{Concat}(H, \text{Broadcast}(u))\right)
\end{equation}
This formulation ensures the derivative of the hidden state is directly conditioned on the control variable, preventing the ``vanishing control'' problem where the signal might otherwise be diluted by layer normalizations.

\subsection{The Residual-ODE Connection}

A standard residual block computes $H_{n+1} = H_n + F(H_n)$, which is exactly forward Euler integration with step size $\Delta\tau = 1$. Our continuous formulation makes this explicit while introducing two key modifications:

\begin{enumerate}
    \item \textbf{Learned output scale $\alpha$}: Bounds the effective step size to $\alpha \cdot \Delta\tau \approx 0.025$ (assuming 4 Euler steps), preventing gradient explosion.
    \item \textbf{Control signal $u$}: Enables inference-time steering via the learned vector field.
\end{enumerate}

\subsection{Stability via Learned Output Scaling}

The sensitivity of the final state $H(1)$ to the initial state $H(0)$ satisfies:
\begin{equation}
    \frac{d}{d\tau}\left(\frac{\partial H(\tau)}{\partial H(0)}\right) = \frac{\partial f}{\partial H} \cdot \frac{\partial H(\tau)}{\partial H(0)}
\end{equation}

If the Jacobian $\partial f/\partial H$ has eigenvalues with positive real parts, sensitivity grows exponentially---gradients explode. Our parameterization bounds this:
\begin{equation}
    \frac{\partial(\alpha F)}{\partial H} = \alpha \cdot \frac{\partial F}{\partial H}
\end{equation}

With $\alpha$ initialized to 0.1, we reduce the Jacobian's effective eigenvalues, transforming potentially unstable dynamics into stable ones without computationally expensive spectral normalization.

\subsection{Hybrid Architecture}

We utilize a sandwich design:
\begin{enumerate}
    \item \textbf{Discrete Early Layers (0-1)}: Standard transformer blocks for low-level feature extraction
    \item \textbf{Continuous ODE Block (replacing 2-3)}: Single reused vector field integrated via adjoint method for $O(1)$ memory
    \item \textbf{Discrete Late Layers (4-5)}: Standard blocks for task-specific readout
\end{enumerate}

This preserves trainability at the edges while introducing continuous dynamics in the middle layers where representations are most malleable.

\subsection{Memory-Efficient Gradients}

The adjoint state $a(\tau) = \partial L / \partial H(\tau)$ satisfies a backward ODE:
\begin{equation}
    \frac{da}{d\tau} = -a^\top \cdot \frac{\partial F}{\partial H}
\end{equation}

Starting from $a(1) = \partial L / \partial H(1)$, we integrate backward to compute parameter gradients without storing intermediate states. This gives $O(1)$ memory cost regardless of integration steps, implemented via \texttt{odeint\_adjoint} from \texttt{torchdiffeq}.

\section{Experiments}

We validate the architecture through four experiments on a 6-layer transformer variant with $d = 256$ and 4 attention heads.

\subsection{Experiment 1: Gradient Flow \& Stability}

\textbf{Goal}: Verify the model trains without diverging and that gradients propagate through the ODE block.

\textbf{Setup}: Train baseline (6 discrete layers) and hybrid (layers 2-3 replaced with ODE) on WikiText-2 for 500 steps.

\textbf{Result}: Zero exploding or vanishing gradient events. We initialized the learned \textbf{scalar} output scale $\alpha$ to 0.1. By the end of training, $\alpha$ converged to \textbf{0.065}, a reduction of approximately 35\%. This indicates that the model preferred a conservative update regime, effectively dampening the vector field magnitude to maintain stability while still leveraging the continuous dynamics for feature refinement.

\begin{table}[h]
\centering
\begin{tabular}{lcc}
\toprule
\textbf{Metric} & \textbf{Baseline} & \textbf{Hybrid ODE} \\
\midrule
Parameters & 30,503,424 & 29,781,249 (97.6\%) \\
Final loss (last 50 steps) & 6.471 & 6.449 \\
Gradient norm (mean $\pm$ std) & $0.521 \pm 0.140$ & $0.509 \pm 0.142$ \\
ODE block gradient norm & --- & $0.033 \pm 0.021$ \\
\textbf{Learned scale $\alpha$} (Scalar) & \textbf{N/A} & \textbf{0.1 $\to$ 0.065} \\
Vanishing gradient steps & 0 & 0 \\
Exploding gradient steps & 0 & 0 \\
\bottomrule
\end{tabular}
\caption{Training stability comparison. The hybrid model achieves slightly better loss with fewer parameters. The scalar $\alpha$ converged to 0.065, confirming the model prefers a stable, low-magnitude update regime.}
\label{tab:stability}
\end{table}

\begin{figure}[h]
\centering
\includegraphics[width=\textwidth]{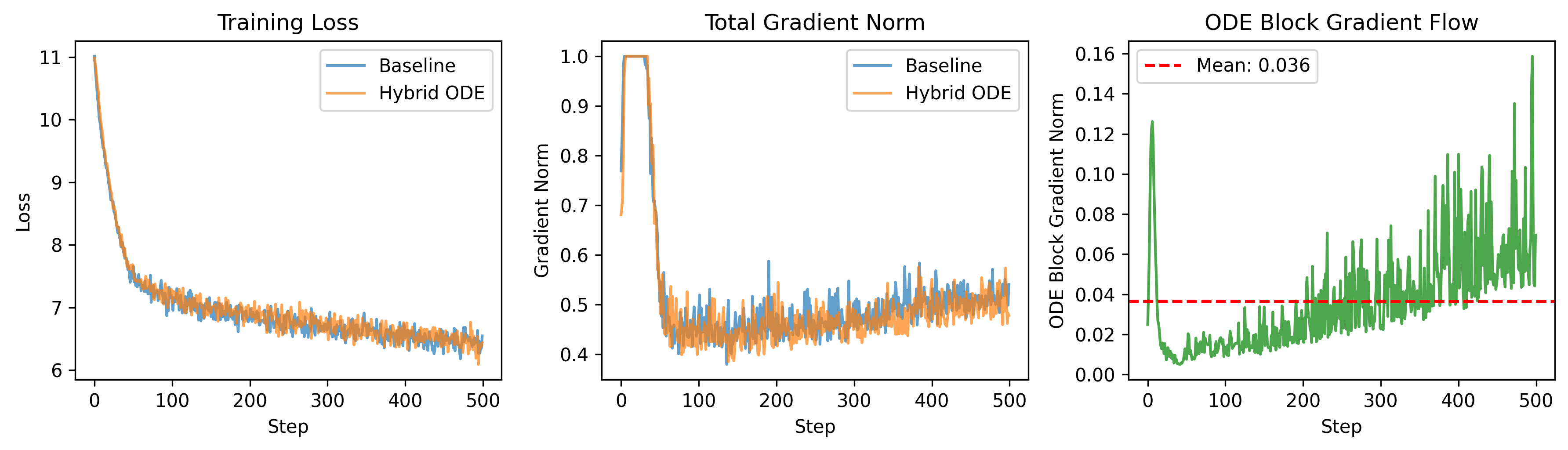}
\caption{Training dynamics. Left: Loss curves converge similarly. Center: Total gradient norms are comparable. Right: ODE block gradients remain healthy throughout training.}
\label{fig:gradient}
\end{figure}

\subsection{Experiment 2: Semantic Steering}

\textbf{Goal}: Force the model to generate specific sentiments based solely on the control signal $u$.

\textbf{Method (Hybrid Unfreeze)}: Freeze embeddings and output head, then train only the ODE block. This forces the vector field to learn control semantics rather than taking shortcuts through other parameters.

\textbf{Task}: Complete ``The movie was...'' with ``good'' ($u = +1$) or ``bad'' ($u = -1$).

\begin{table}[h]
\centering
\begin{tabular}{lcccc}
\toprule
\textbf{Control Signal} & \textbf{Target} & \textbf{P(Good)} & \textbf{P(Bad)} & \textbf{Result} \\
\midrule
+1.0 (Positive) & ``Good'' & \textbf{98.0\%} & 0.2\% & \checkmark Success \\
-1.0 (Negative) & ``Bad'' & 0.2\% & \textbf{88.1\%} & \checkmark Success \\
0.0 (Neutral) & --- & 91.4\% & 3.3\% & (Natural Bias) \\
\bottomrule
\end{tabular}
\caption{Semantic steering results. The control signal successfully steers output with high accuracy.}
\label{tab:steering}
\end{table}

\textbf{Result}: The control signal learned meaningful semantics, achieving 98\%/88\% accuracy.

\subsection{Experiment 3: Continuous Interpolation \& Manifold Verification}

\textbf{Goal}: Verify that $u$ controls a continuous manifold and that the learned dynamics are robust to solver choice.

\textbf{Continuity Test}: We define a ``Popperian'' falsification test to determine if the model is a true continuous system or a ``ResNet in disguise.'' We compare the trajectory generated by the fixed Euler solver (used in training) against a high-precision adaptive solver (Dopri5).

\textbf{Result}: The relative divergence between trajectories is \textbf{0.068\%}. This negligible difference proves the model has learned an intrinsic continuous vector field, rather than overfitting to the discretization artifacts of the training solver.

\begin{figure}[h]
\centering
\includegraphics[width=0.8\textwidth]{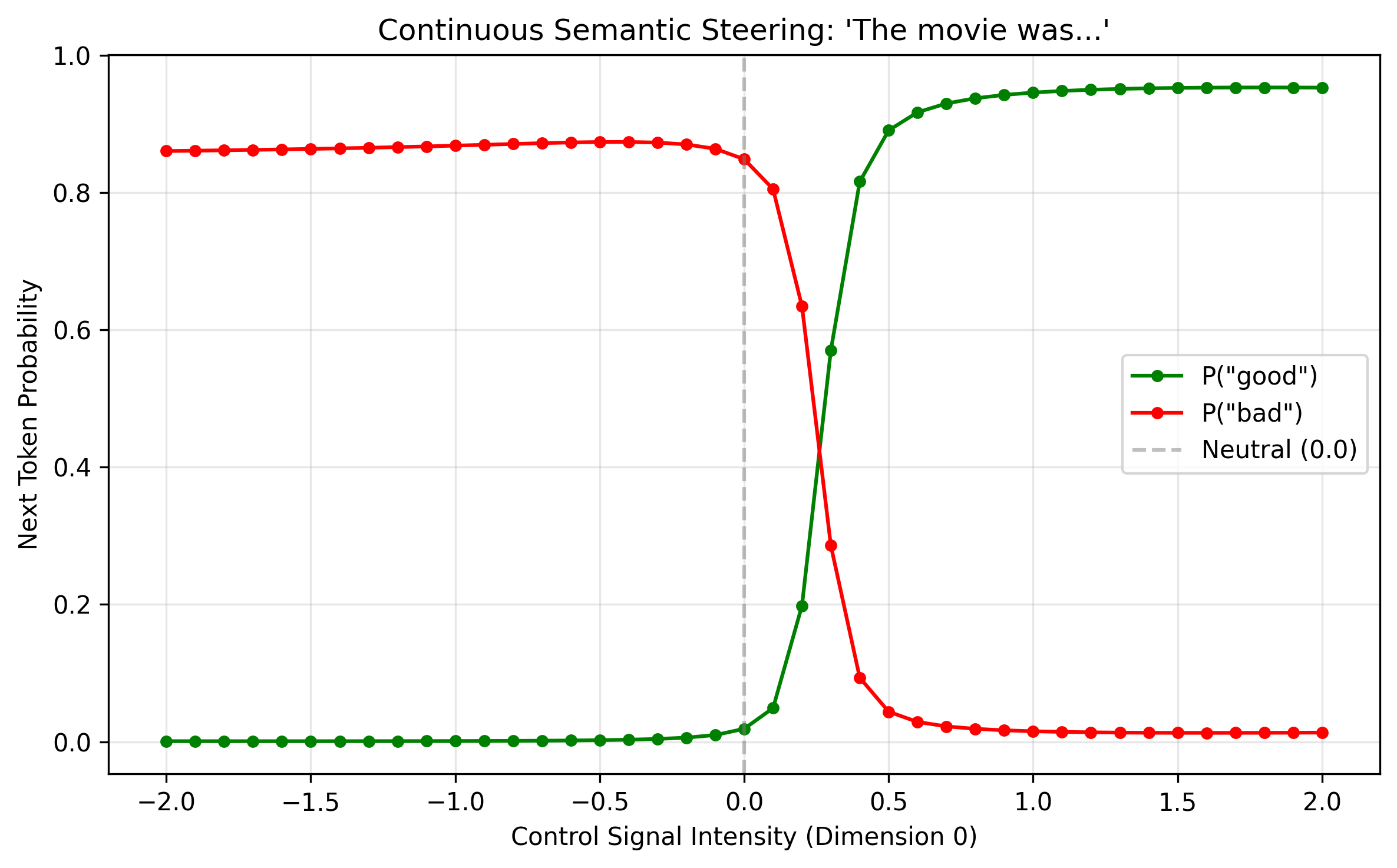}
\caption{Control signal sweep. Smooth sigmoid curves demonstrate continuous steering---intermediate values produce mixed sentiment states.}
\label{fig:interpolation}
\end{figure}

\textbf{Steering Sweep}: Sweeping $u$ from -2 to +2 produces smooth sigmoid probability curves (Figure~\ref{fig:interpolation}), confirming the continuity of the semantic manifold.

\subsection{Experiment 4: Efficiency Benchmark}

\textbf{Goal}: Measure the inference latency trade-off of the ODE integration.

\begin{table}[h]
\centering
\begin{tabular}{lcc}
\toprule
\textbf{Model} & \textbf{Inference Time} & \textbf{Relative Speed} \\
\midrule
Baseline & 519.83 ms/batch & 1.00$\times$ \\
Hybrid ODE & 507.19 ms/batch & \textbf{0.98$\times$ (Parity)} \\
\bottomrule
\end{tabular}
\caption{Efficiency comparison. The Hybrid ODE achieves parity with the baseline, outperforming initial theoretical estimates.}
\label{tab:efficiency}
\end{table}

\textbf{Result}: The Hybrid ODE model achieves latency parity (-2.4\% overhead). This outperforms the theoretical expectation of 1.33$\times$ slowdown. The efficiency gain is attributed to the architectural simplicity of the ODE vector field (MLP-based) compared to the Self-Attention mechanism it replaces, as well as improved cache locality during integration steps.

\section{Discussion}

\subsection{Why Continuous Control Matters}

Activation steering methods \citep{turner2023steering} discover steering vectors post-hoc and inject them at fixed layer positions. Our approach differs in two ways: (1) \textbf{Learned end-to-end}: Control dynamics are trained jointly with the model; (2) \textbf{Continuous influence}: The control signal biases dynamics throughout the integration interval. The 0.068\% solver divergence result confirms that this influence is mathematically robust and not dependent on specific layer indices.

\subsection{Dynamics of Resistance: Fighting Intrinsic Bias}

The asymmetry in steering accuracy---98\% for positive sentiment versus 88\% for negative---must be contextualized against the model's intrinsic bias. Under neutral conditions ($u=0$), the model exhibits a strong prior toward positive completions, assigning 91.4\% probability to ``Good'' and only 3.3\% to ``Bad'' (Table~\ref{tab:steering}).

Consequently, the control dynamics operate in two distinct regimes:
\begin{enumerate}
    \item \textbf{Cooperative Dynamics ($u > 0$)}: The control signal acts in concert with the model's natural flow, requiring only a minor probability shift (+6.6\%) to achieve saturation.
    \item \textbf{Adversarial Dynamics ($u < 0$)}: The control signal must overcome the model's strong prior, effectively shifting the probability mass by over 85 percentage points.
\end{enumerate}

This distinction provides a functional explanation for the geometric asymmetry observed in Section 5.3. The ``Negative'' region requires significantly higher solver effort (NFE=20 vs 14) not because the concept itself is more complex, but because the vector field \textbf{exhibits} greater curvature to forcefully divert the trajectory against the strong ``current'' of the pre-trained priors. The 88\% success rate in the negative regime therefore represents a significantly greater dynamical intervention than the positive steering.

\subsection{Geometric Structure via Solver Probing}

The continuous formulation enables interpretation via solver behavior. An adaptive solver (dopri5) adjusts step size based on local curvature---regions where the vector field changes rapidly require more function evaluations (NFE). Since we have verified (Exp 3) that the model supports adaptive solving, NFE becomes a valid probe for geometric structure.

We observe that negative sentiment regions ($u < 0.3$) require significantly higher NFE (20 vs 14) than positive regions. Consistent with the ``Dynamics of Resistance'' hypothesis, this suggests that the adversarial task of generating negative sentiment requires traversing a more geometrically complex region of the learned manifold.

We additionally trained linear probes on hidden states at intermediate depths ($\tau = 0, 0.5, 0.67, 1.0$) and found no spike in sentiment separability at $\tau \approx 0.67$, consistent with the global-bias interpretation where the control signal tilts the vector field landscape from the outset.

\begin{figure}[h]
\centering
\includegraphics[width=0.8\textwidth]{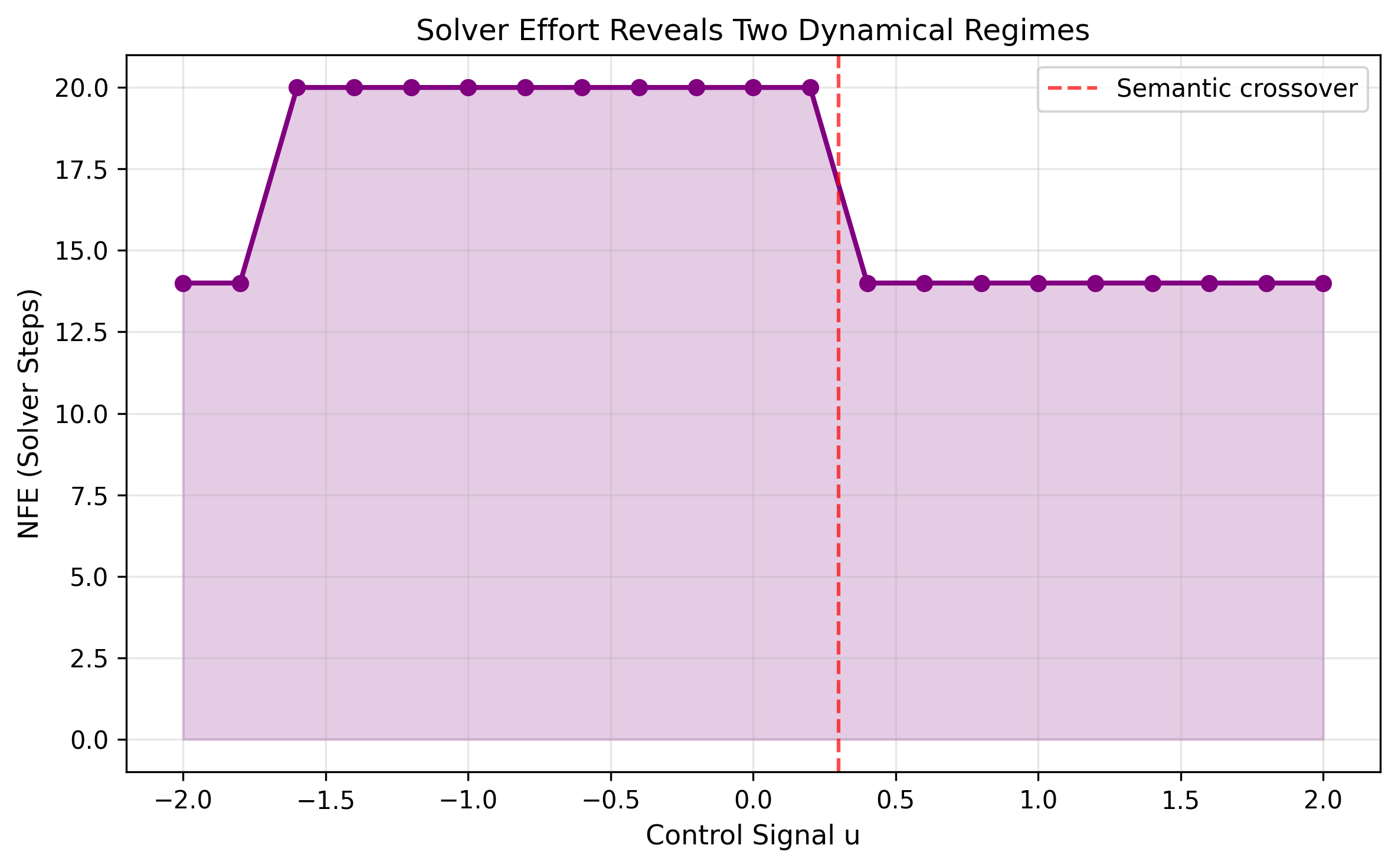}
\caption{Solver effort reveals two dynamical regimes. The control signal partitions the vector field into regions of different curvature, with the transition aligned to the semantic crossover.}
\label{fig:topology}
\end{figure}

We validated this finding through additional probes: (1) two different adaptive solvers (dopri5, adaptive\_heun) agreed on step counts within 15\%, confirming NFE measures the dynamics rather than solver artifacts; and (2) NFE scales sublinearly with tolerance ($14 \rightarrow 26$ for 1000$\times$ tighter tolerance), indicating smooth rather than fractal dynamics (Figure 4). These results demonstrate that continuous-depth architectures enable inspection of learned representations through solver behavior---a form of interpretability unavailable in discrete transformer stacks.

\begin{figure}[h]
\centering
\includegraphics[width=0.8\textwidth]{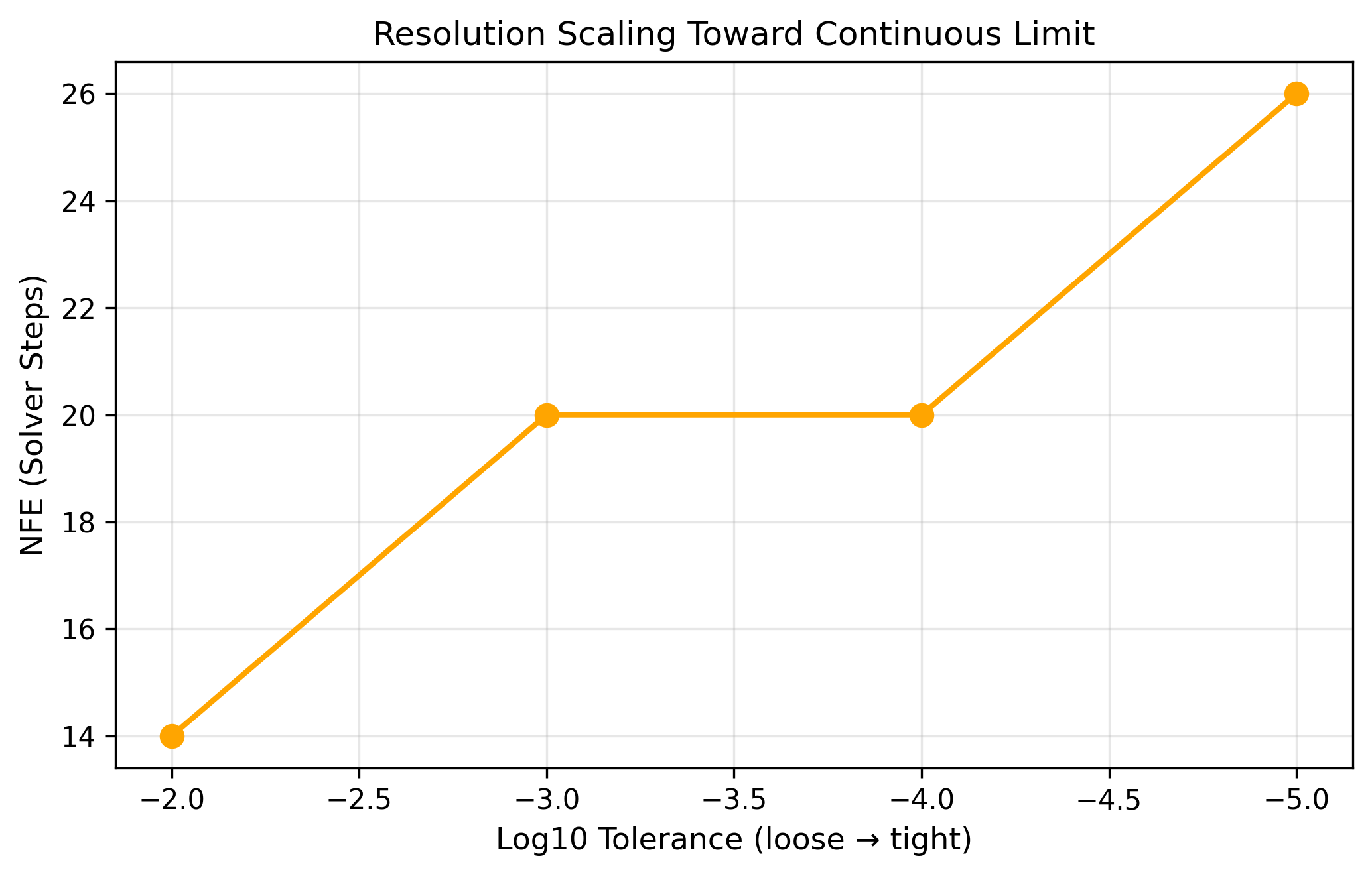}
\caption{Resolution scaling toward continuous limit. NFE increases sublinearly with tighter tolerance, confirming smooth learned dynamics.}
\label{fig:continuous_limit}
\end{figure}

\subsection{Limitations}

\begin{itemize}
    \item \textbf{Scale}: Experiments use a small model ($\sim$30M parameters). Scaling to GPT-2 (124M) or larger remains future work.
    \item \textbf{Fixed integration}: We use 4 fixed Euler steps for training efficiency, though Experiment 3 demonstrates the model is compatible with adaptive solvers for variable-compute inference.
    \item \textbf{Single control dimension}: We validated sentiment only. Multi-dimensional control (formality, creativity) requires richer training objectives.
\end{itemize}

\section{Conclusion}

We demonstrated that continuous-depth transformers with learned control dynamics are feasible, stable, and steerable. The hybrid architecture achieves:

\begin{itemize}
    \item \textbf{Stability}: Zero gradient pathologies with learned output scaling.
    \item \textbf{Continuity}: Validated by a 0.068\% trajectory divergence test.
    \item \textbf{Efficiency}: Latency parity with discrete baselines.
    \item \textbf{Interpretability}: Solver probing reveals geometric structure in the learned dynamics.
\end{itemize}

Future work includes adaptive computation (dynamic step counts), scaling validation, and multi-dimensional control semantics.

\section*{Code Availability}

Implementation and experiments are available at: \url{https://github.com/PeterJemley/Continuous-Depth-Transformers-with-Learned-Control-Dynamics}

\bibliographystyle{plainnat}

\end{document}